\documentclass{article}
\usepackage{ijcai20}

\usepackage{times}
\usepackage{soul}
\usepackage{url}
\usepackage[hidelinks]{hyperref}
\usepackage[utf8]{inputenc}
\usepackage[small]{caption}
\usepackage{graphicx}
\usepackage{amsmath}
\usepackage{amsthm}
\usepackage{booktabs}
\usepackage{algorithm}
\usepackage{algorithmic}
\usepackage{multirow}
\usepackage{amsfonts}
\usepackage{epsfig}
\usepackage{mathtools}
\usepackage[caption=false]{subfig}
\usepackage{amssymb}
\usepackage{array}
\usepackage{xcolor}

\title{Arid: A New Dataset for Recognizing Action in the Dark}

\author{
	Yuecong Xu$^1$
	\and
	Jianfei Yang$^1$\and
	Haozhi Cao$^1$\and
	Kezhi Mao$^1$\and
	Jianxiong Yin$^2$\And
	Simon See$^2$
	\affiliations
	$^1$School of Electrical and Electronic Engineering, Nanyang Technological University\\
	$^2$NVIDIA AI Tech Centre
	\emails
	\{xuyu0014, yang0478, haozhi001, ekzmao\}@ntu.edu.sg,
	\{jianxiongy, ssee\}@nvidia.com,
}

\newcolumntype{C}[1]{>{\centering\arraybackslash}m{#1}}

\begin{document}
	
	\maketitle
	
	%%%%%% Abstract STARTS HERE %%%%%%%%%
	\begin{abstract}
		The task of action recognition in dark videos is useful in various scenarios, e.g., night surveillance and self-driving at night. Though progress has been made in the action recognition task for videos in normal illumination, few have studied action recognition in the dark. This is partly due to the lack of sufficient datasets for such a task. In this paper, we explored the task of action recognition in dark videos. We bridge the gap of the lack of data for this task by collecting a new dataset: the Action Recognition in the Dark (ARID) dataset. It consists of over 3,780 video clips with 11 action categories. To the best of our knowledge, it is the first dataset focused on human actions in dark videos. To gain further understandings of our ARID dataset, we analyze the ARID dataset in detail and exhibited its necessity over synthetic dark videos. Additionally, we benchmarked the performance of several current action recognition models on our dataset and explored potential methods for increasing their performances. Our results show that current action recognition models and frame enhancement methods may not be effective solutions for the task of action recognition in dark videos.
	\end{abstract}
	
	%%%%%%%%%% BODY TEXT STARTS HERE %%%%%%%%%%
	\section{Introduction}
	\label{section:introduction}
	
	Thanks to the increasing application of automatic action recognition in various fields, such as surveillance~\cite{zou2019wifi,chen2010real,ullah2021efficient} and smart homes~\cite{fahad2015integration,feng2017smart,yang2018device}, action recognition tasks have received considerable attention in recent years. Although much progress has been made, current research still mostly focuses on videos shot under normal illumination. This is partly due to the fact that current benchmark datasets for action recognition~\cite{kuehne2011hmdb,soomro2012ucf101,carreira2017quo} are normally collected from web videos, which are shot mostly under normal illumination. Yet videos under normal illumination conditions are not available in many cases, such as night surveillance~\cite{hatakeyama2008detection,soumya2017recolorizing}, and self-driving at night~\cite{maqueda2018event}. It is true that additional sensors, such as infrared or thermal imaging sensors, could be utilized for recognizing actions in the dark. However, such sensors are of high cost and the deployment of such sensors on a large scale may not be practical. Hence we focus on action recognition in the dark without the need for additional sensors. To this end, we collected a new dataset: Action Recognition In the Dark (ARID) dataset, dedicated to the task of action recognition in dark videos. To the best of our knowledge, it is the first dataset focused on human actions in the dark.
	
	Currently, there already exists a large number of videos in various datasets, shot under normal illumination. Intuitively, we may make use of these videos to create synthetic dark videos. In this paper, we will show the distinct characteristics of real dark videos that cannot be replicated by synthetic dark videos through detailed analysis and comparison. This implies that a real dark video dataset is necessary for the task of action recognition in the dark.
	
	Recently, neural networks, especially convolutional neural network (CNN) based solutions have proven to be effective for various computer vision tasks, including the action recognition task. For action recognition, state-of-the-art results on previous action recognition datasets are mostly achieved through either two-stream networks~\cite{simonyan2014two,dai2020human} or 3D-CNN-based networks~\cite{tran2015learning,ullah2021efficient}. To gain further understanding of the challenges faced with action recognition in dark videos, we analyze how dark videos affect current action recognition models. Additionally, we explore potential methods for substantial improvements in action recognition accuracy utilizing current action recognition models.
	
	In summary, we explored the task of action recognition in dark videos. The contribution of this work is three-fold and is summarized as follows:
	\begin{itemize}
		\item We develop a new dataset, the Action Recognition in the Dark (ARID) dataset, dedicated to the task of recognizing actions in dark videos, which, to the best of our knowledge, is the first dataset focused on human actions in dark videos.
		\item We discover the distinct characteristics of real dark videos through statistical and visual analysis and comparison with synthetic dark videos.
		\item We benchmark the performance of various action recognition models on our dataset while exploring potential methods to improve action recognition accuracy with current models, and reveals challenges in the task of action recognition in dark videos.
	\end{itemize}
	
	The rest of this paper is organized as follows: related works of datasets for action recognition tasks and visual tasks in the dark environments are discussed in Section~\ref{section:related}. In Section~\ref{section:ARID}, we introduce our proposed Action Recognition In the Dark (ARID) dataset in detail. After that, we benchmark current action recognition models on our dataset, with a thorough analysis of our dataset and its challenges in Section~\ref{section:exp}. Finally, we conclude the paper and propose our future work in Section~\ref{section:conclusion}.

	%------------------------------------------------------------------------
	\section{Related Works}
	\label{section:related}
	%%%%%%%%%% 需要补充包括ASC文章相关 %%%%%%%%%% 
	
	\subsection{Action Recognition Datasets}
	\label{section:related:ar-datasets}
	There are a number of benchmark datasets in the action recognition domain. Earlier datasets, such as KTH~\cite{schuldt2004recognizing}, Weizmann~\cite{gorelick2007actions} and IXMAMS~\cite{weinland2007action}, contain relatively small number of action classes. These datasets are mostly shot offline without using publicly available online videos. With the rapidly increased performance of proposed models on these smaller datasets, larger and more challenging datasets are introduced subsequently. These includes HMDB51~\cite{kuehne2011hmdb}, UCF101~\cite{soomro2012ucf101}, Activity-Net~\cite{caba2015activitynet} and Kinetics~\cite{carreira2017quo}. Compared to previous datasets, they contain a larger variety of actions, with more videos for each action than previous datasets. Particularly, the Kinetics dataset~\cite{carreira2017quo}, with 400 action classes and more than 160,000 clips in total, becomes the primary choice in action recognition studies. The introduction of these larger datasets help pushed the boundaries of the task of action recognition, and lead to the introduction of more sophisticated models. However, although these datasets involve an abundant scale of actions, these actions are mostly collected from web videos due to the ease of obtaining web videos through mass downloading. These web videos are mostly recorded under normal illumination, which limits the capability of current action recognition models to videos with normal illumination. Hence, to study the action recognition performance in dark videos, we collected a new video dataset dedicated to videos shot in the dark.
	
	\subsection{Dark Visual Datasets}
	\label{related:dark-datasets}
	Recently, there has been a rise of research interest with regards to computer vision tasks in a dark environment, such as face recognition in the dark~\cite{han2013comparative,shim2008subspace,chen2006total}. The research for dark environment visual tasks is partly supported by the various dark visual datasets introduced. Among these, most datasets focused on image enhancement and denoising tasks, where the goal is to visually enhance dark images for a clearer view. These include LOL Dataset~\cite{chen2018retinex}, ReNOIR~\cite{anaya2018renoir}, ExDARK~\cite{loh2019getting} and SID~\cite{chen2018learning}. Images in these datasets are shot in a low illumination environment, where it is difficult for the human naked eye to observe and identify objects. More recently, such an enhancement task has been expanded to the video domain, where the goal is to process dark videos for clearer video frames with higher visibility. These include DRV~\cite{chen2019seeing} and SMOID~\cite{jiang2019learning} datasets. Although both datasets contain dark videos, their focus is more on enhancing the visibility of video frames. Clearer video frames may break the original pixel distribution, thus there is no guarantee that clearer videos would result in higher action recognition accuracy. Furthermore, the scenes included in these video datasets are randomly shot and may not include specific human actions, thus are not suitable for the task of action recognition. In contrast, our ARID dataset focuses on classifying different human actions in dark videos.

	%------------------------------------------------------------------------
	\section{Action Recognition In the Dark Dataset}
	\label{section:ARID}
	
	\begin{figure}[!t]
		\centering
		\includegraphics[width=0.75\linewidth]{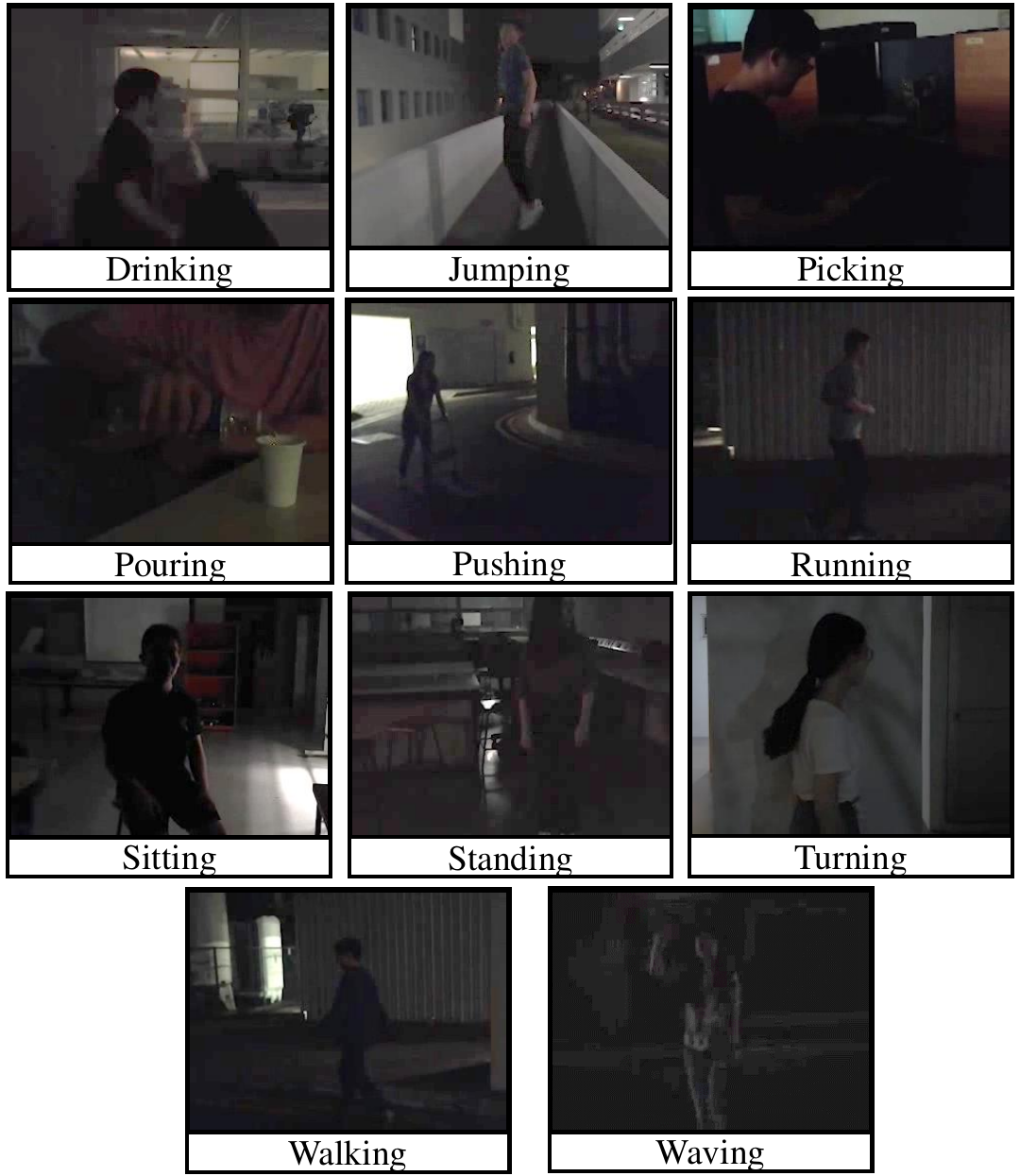}
		\caption{Sample frames for each of the 11 action classes of the ARID dataset. All samples are manually tuned brighter for display purposes.}
		\label{figure:ARID-1-sample}
	\end{figure}
	
	Although a small number of videos taken in the dark do exist in current action recognition benchmark datasets, such as Kinetics~\cite{carreira2017quo} and HMDB51~\cite{kuehne2011hmdb}, the task of human action recognition in dark environments has rarely been studied. This is partly due to the very low proportion of dark videos in current benchmark datasets, and a lack of datasets dedicated to action analysis in the dark. To bridge the gap in the lack of dark video data, we introduce a new Action Recognition In the Dark (ARID) dataset. In this session, we take an overview of the dataset from three perspectives: the action classes, the process of data collection as well as some basic statistics of our ARID dataset.
	
	\subsection{Action Classes}
	\label{section:ARID:classes}
	The ARID dataset includes a total of 11 common human action classes. The list of action classes can be categorized into two types: \textit{Singular Person Actions}, which includes jumping, running, turning, walking, and waving; and \textit{Person Actions with Objects}, which includes drinking, picking, pouring, pushing, sitting, and standing. Figure~\ref{figure:ARID-1-sample} shows the sample frames for each of the 11 action classes in the ARID dataset.
	
	\subsection{Data Collection}
	\label{section:ARID:collection}
	The video clips in the ARID dataset are collected using 3 different commercial cameras available in the market. The clips are shot strictly during night hours. All clips are collected from a total of 11 volunteers, among which 8 males and 3 females. We collected the clips in 9 outdoor scenes and 9 indoor scenes, such as carparks, corridors and playing fields for outdoor scenes, and classrooms and laboratories for indoor scenes. The lighting condition of each scene is different, with no direct light shot on the actor in almost all videos. In many cases, it is challenging even for the naked eye to recognize the human action without tuning the raw video clips.
	
	\begin{figure}[t]
		\centering
		\includegraphics[width=0.75\linewidth]{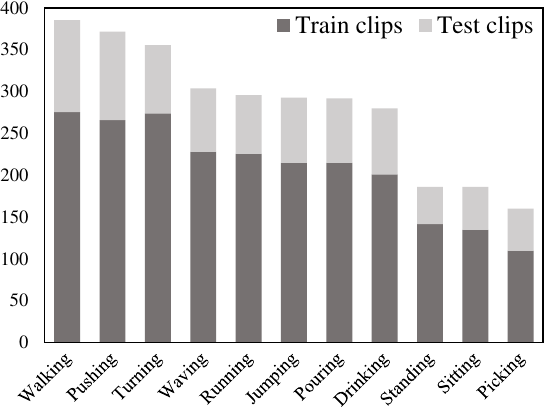}
		\caption{The distribution of clips among all action classes in ARID. The dark grey and light grey bars indicate the number of clips in the train and test partitions.}
		\label{figure:ARID-2-dist}
	\end{figure}
	
	\subsection{Basic Statistics}
	\label{section:ARID:statistics}
	The ARID dataset contains a total of 3,784 video clips, with each class containing at least 110 clips. The clips of a single action class are divided into 12-18 groups with each group containing no less than 7 clips. The clips in the same group share some similar features, such as being shot under similar lighting conditions or shot with the same actor. Figure~\ref{figure:ARID-2-dist} shows the distribution of clips among all the classes.
	
	The training and testing sets are partitioned by splitting the clip groups, with $70\%$ of the groups in the training partition, and the remaining $30\%$ of the groups in the testing partition. We selected three training/testing splits, such that each group would have an equal chance to be present in either the training partition or the testing partition.
	
	The video clips are fixed to a frame rate of 30 FPS with a resolution of $320\times240$. The minimum clip length is 1.2 seconds with 36 frames, and the duration of the whole dataset is 8,721 seconds. The videos are saved in \texttt{.avi} format and are compressed using the \textit{DivX} codec.

	%------------------------------------------------------------------------
	\section{Experiments and Discussions}
	\label{section:exp}
	
	In this section, we gain further understandings of our proposed ARID dataset through a detailed analysis of the dataset. The main objectives are two-fold: 1) validate the necessity of a video dataset collected in the real dark environment and 2) provide a benchmark for current action recognition models and methods while revealing the challenges with regards to the task of action recognition in dark videos. In the following, we first introduce the experiment settings along with the construction of a synthetic dark video dataset. We then introduce methods used to enhance dark video frames in ARID in an effort to improve action recognition accuracy. We then analyze our introduced ARID dataset in detail through three perspectives: statistical and visual analysis of ARID, analysis of the ARID classification results, and visualization of extracted features from videos in ARID.
	
	\subsection{Experimental Settings}
	\label{section:exp:settings}
	To obtain the action recognition results on our ARID dataset, we utilize both two-stream models and 3D-CNN-based models with PyTorch\cite{paszke2019pytorch}. For our experiments, the inputs to all 3D-CNN-based models are sequences of 16 sampled frames, with each frame resized to $224\times224$. The inputs to the spatial stream of our two-stream models are RGB sampled frames, resized to $224\times224$. Whereas the inputs to the temporal stream are stacks of optical flow pre-computed on both $x$ and $y$ axis and resized to the same size as the input for the spatial stream, i.e.\ $224\times224$. To accelerate training, we utilize the pretrained models pretrained on the Kinetics~\cite{carreira2017quo} or ImageNet~\cite{deng2009imagenet} dataset when available. Due to the constraints in computation power, a unified batch size of 16 is applied to all 3D-CNN model experiments, while a unified batch size of 32 is applied to all two-stream model experiments. The action recognition results are reported as the average top-1 and average top-5 accuracies over the three splits.
	
	Compared to collecting a new dataset for the dark environment, it is more intuitive to obtain "dark`` videos through synthesizing dark videos from current publicly available datasets, which mainly consist of videos shot under normal illumination. To showcase the necessity of a real dark video dataset, we compare a synthetic dark video dataset with our introduced ARID. More specifically, the synthetic dark video dataset is constructed based on the HMDB51~\cite{kuehne2011hmdb} dataset, denoted as HMDB51-dark. We synthesize dark videos by gamma intensity correction formulated as:
	\begin{equation}
		\label{eqn:exp:gamma-darken}
		D(t, x, y) = {I(t, x, y)} ^ {(1/\gamma)}
	\end{equation}
	where $D(t, x, y)$ is the value of the pixel in the synthetic dark video, located at spatial location $(x, y)$ at the $t^{th}$ frame, and $I(t, x, y)$ is the pixel value at the corresponding location in the original video. Both $D(t, x, y)$ and $I(t, x, y)$ are normalized to the range of $[0, 1]$. $\gamma$ is the parameter that controls the degree of "darkness`` in the synthetic dark video, typically in the range of $[0.1, 10]$, where a smaller number would result in lower pixel values, producing darker synthetic videos.
	
	\begin{figure}[!t]
		\centering
		\includegraphics[width=1.0\linewidth]{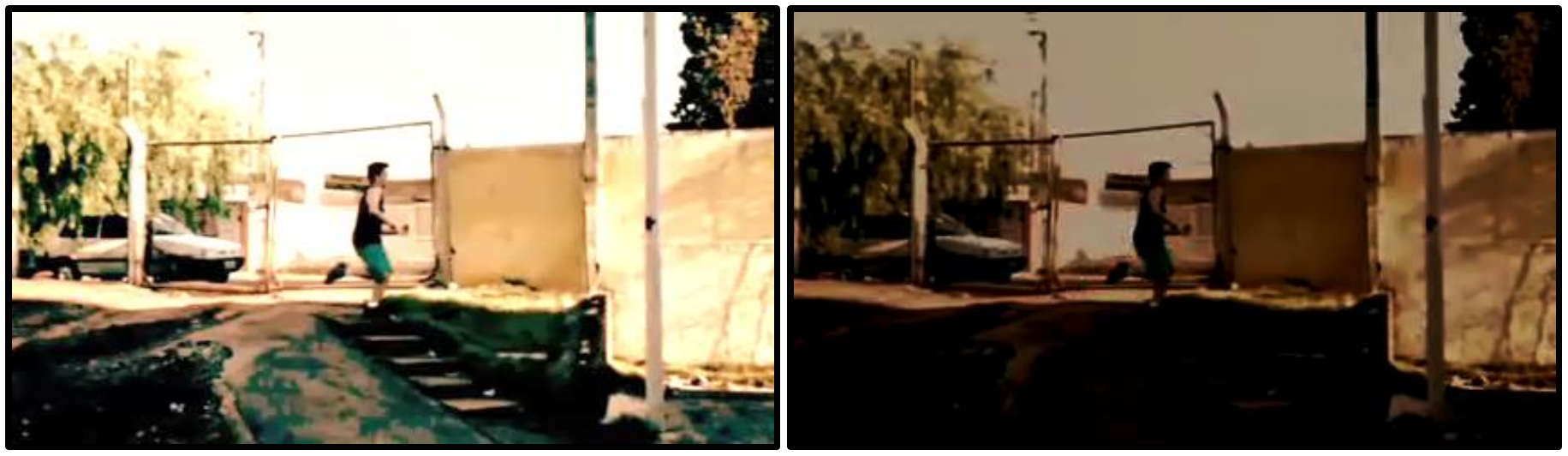}
		\caption{Comparison of a sample frame of normal illumination taken from the video in the HMDB51 dataset (left) and the corresponding frame taken from the synthetic dark video from our HMDB51-dark dataset (right). The frame in the original HMDB51 video has more details, including the background and a clearer contour of the actor. Best viewed in color.}
		\label{figure:exp-3-hmdb_hmdbd}
	\end{figure}
	
	We note that the dark videos collected in our ARID are shot under different illumination conditions. To simulate the differences in illumination within dark videos, we apply different $\gamma$ values when synthesizing dark videos. More specifically, the $\gamma$ value is obtained randomly from a normal distribution $\mathcal{N}(\mu, \sigma^2)$ with the constraint of $\gamma \geq 0.1$. Here the mean $\mu$ is set to 0.2 and the standard deviation $\sigma$ is set to 0.07. Figure~\ref{figure:exp-3-hmdb_hmdbd} shows the comparison of sample frames of videos from the original HMDB51 dataset with that from the corresponding synthetic dark videos.
	
	\subsection{Frame Enhancement Methods}
	\label{section:exp:frame-enhance}
	For humans to better recognize actions in dark videos, an intuitive method is to enhance each dark video frame, such that objects and actions are visually clearer. In this paper, to better understand the effect of dark videos on current action recognition models, we investigate the effect of applying different frame enhancement methods on ARID. Specifically, we applied five common frame enhancement methods: Histogram Equalization (\textbf{HE})~\cite{trahanias1992color}, Gamma Intensity Correction (\textbf{GIC}), \textbf{LIME}~\cite{guo2016lime}, \textbf{BIMEF}~\cite{ying2017new} and \textbf{KinD}~\cite{zhang2019kindling}. Among them, \textbf{HE} and \textbf{GIC} are traditional image enhancement methods. In particular, \textbf{HE} produces higher contrast images, whereas \textbf{GIC} is often used to adjust the luminance of images. Whereas both \textbf{LIME} and \textbf{BIMEF} are based on the Retinex theory~\cite{land1977retinex}, which assumes that images are composed of reflection and illumination. \textbf{LIME} estimates the illumination map of dark images while imposing a structure prior to the initial illumination map, while \textbf{BIMEF} proposes a multi-exposure fusion algorithm. \textbf{KinD} is a more recent deep neural network-based method utilizing a two-stream structure for simultaneous reflectance restoration and illumination adjustment. KinD is implemented with weights pretrained on the LOL Dataset~\cite{chen2018retinex}. The results of applying the above methods to the ARID dataset are denoted as ARID-HE, ARID-GIC, ARID-LIME, ARID-BIMEF, and ARID-KinD respectively. The \textbf{GIC}is also applied to the synthetic dark dataset HMDB51-dark, whose result is denoted as HMDB51-dark-GIC.
	
	\subsection{Statistical and Visual Analysis of ARID}
	\label{section:exp:stat-analysis}
\begin{figure*}[!t]
		\centering
		\includegraphics[width=.85\linewidth]{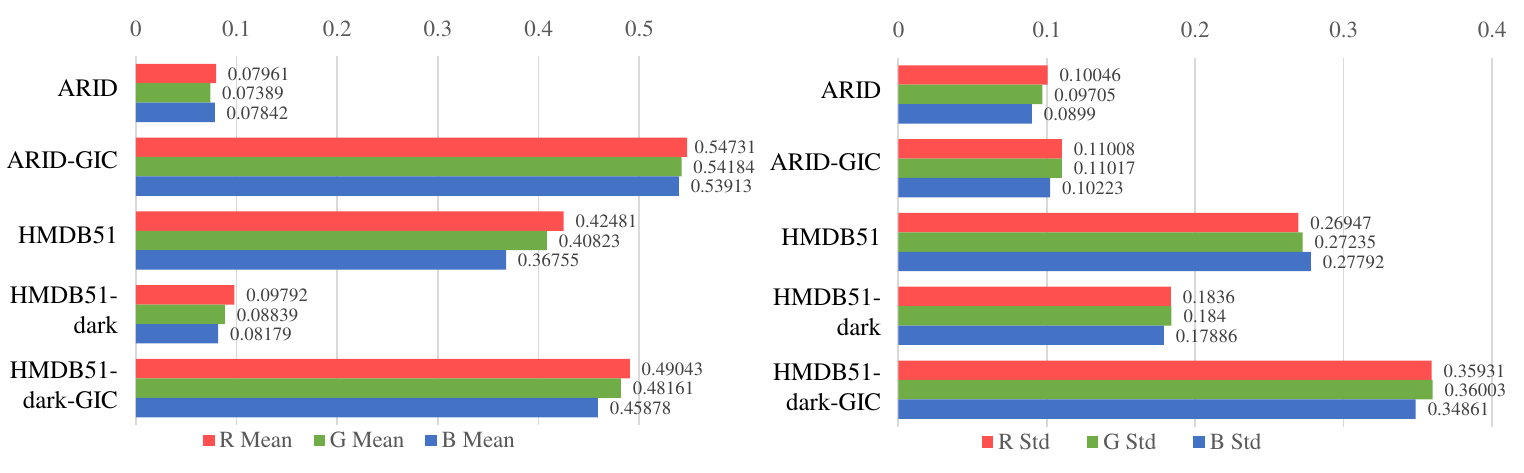}
		\caption{Bar charts of the RGB mean (left) and standard deviation (right) values for various datasets, including ARID and its \textbf{GIC} enhanced output ARID-GIC, HMDB51 and the synthetic dark dataset HMDB51-dark, as well as the \textbf{GIC} enhanced output of the synthetic dart dataset, HMDB51-dark-GIC. All values are normalized to the range of [0.0. 1.0]. Best viewed in color.}
		\label{figure:exp-4-mean_stds}
	\end{figure*}

\begin{figure}[!t]
		\centering
		\includegraphics[width=1.0\linewidth]{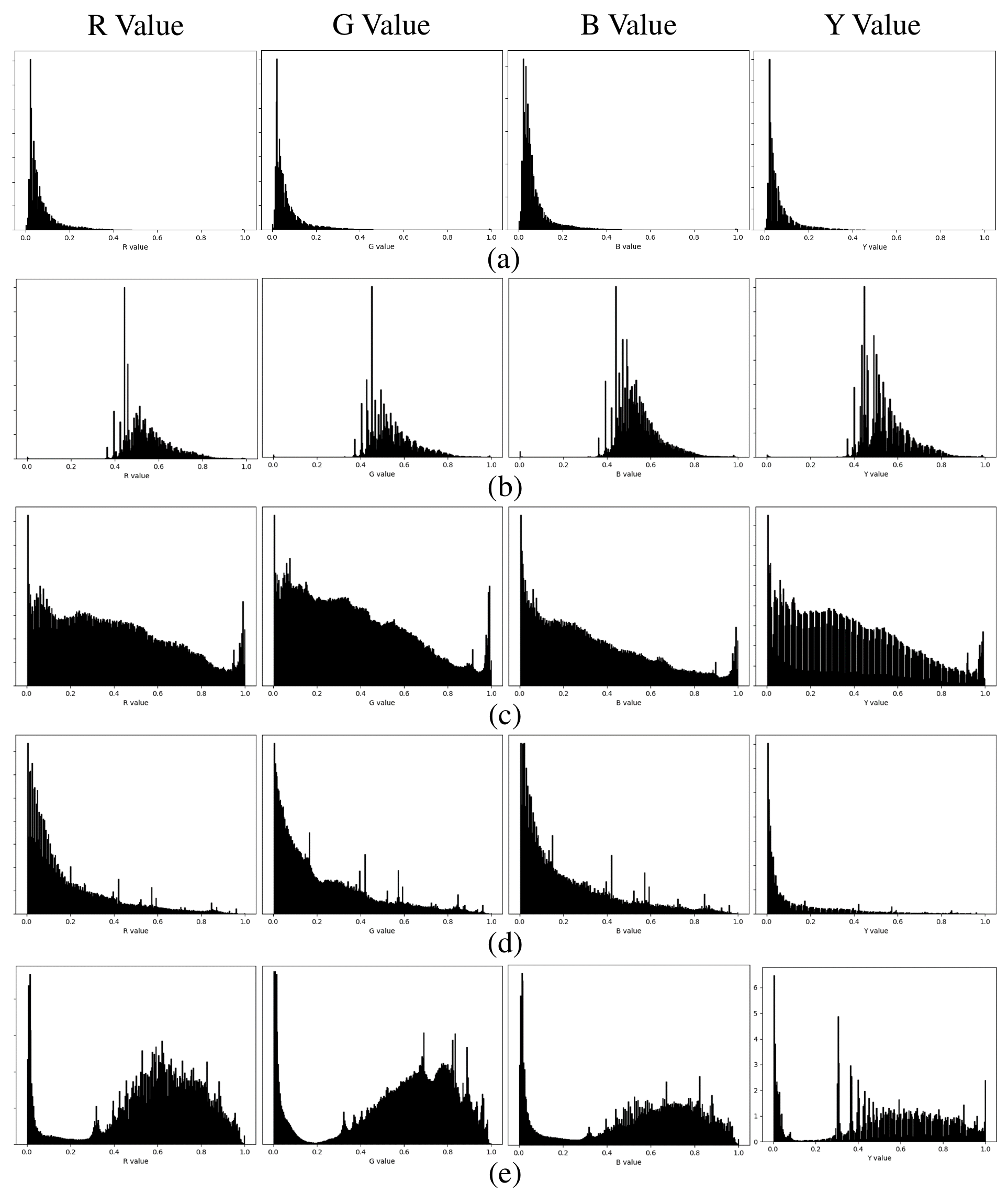}
		\caption{Histograms for RGB and Y values of (from top to bottom): (a) ARID, (b) ARID-GIC, (c) HMDB51, (d) HMDB51-dark and (e) HMDB51-dark-GIC. All values are normalized to the range of [0.0. 1.0]. Best viewed zoomed in.}
		\label{figure:exp-5-rgby_histograms}
	\end{figure}

	To better understand the necessity of real dark videos, we compute and compare the statistics of the ARID dataset with the HMDB51 dataset as well as the synthetic HMDB51-dark dataset. Figure~\ref{figure:exp-5-rgby_histograms} displays the RGB values and Y value histograms of datasets ARID, ARID-GIC, HMDB51, HMDB51-dark and HMDB51-dark-GIC. We also provide the bar charts of the mean value and standard deviation value of the above datasets as shown in Figure~\ref{figure:exp-4-mean_stds}. The gamma values for obtaining both ARID-GIC and HMDB51-dark-GIC are both set to $\gamma=5$.
	
\begin{figure}[!t]
		\centering
		\includegraphics[width=.9\linewidth]{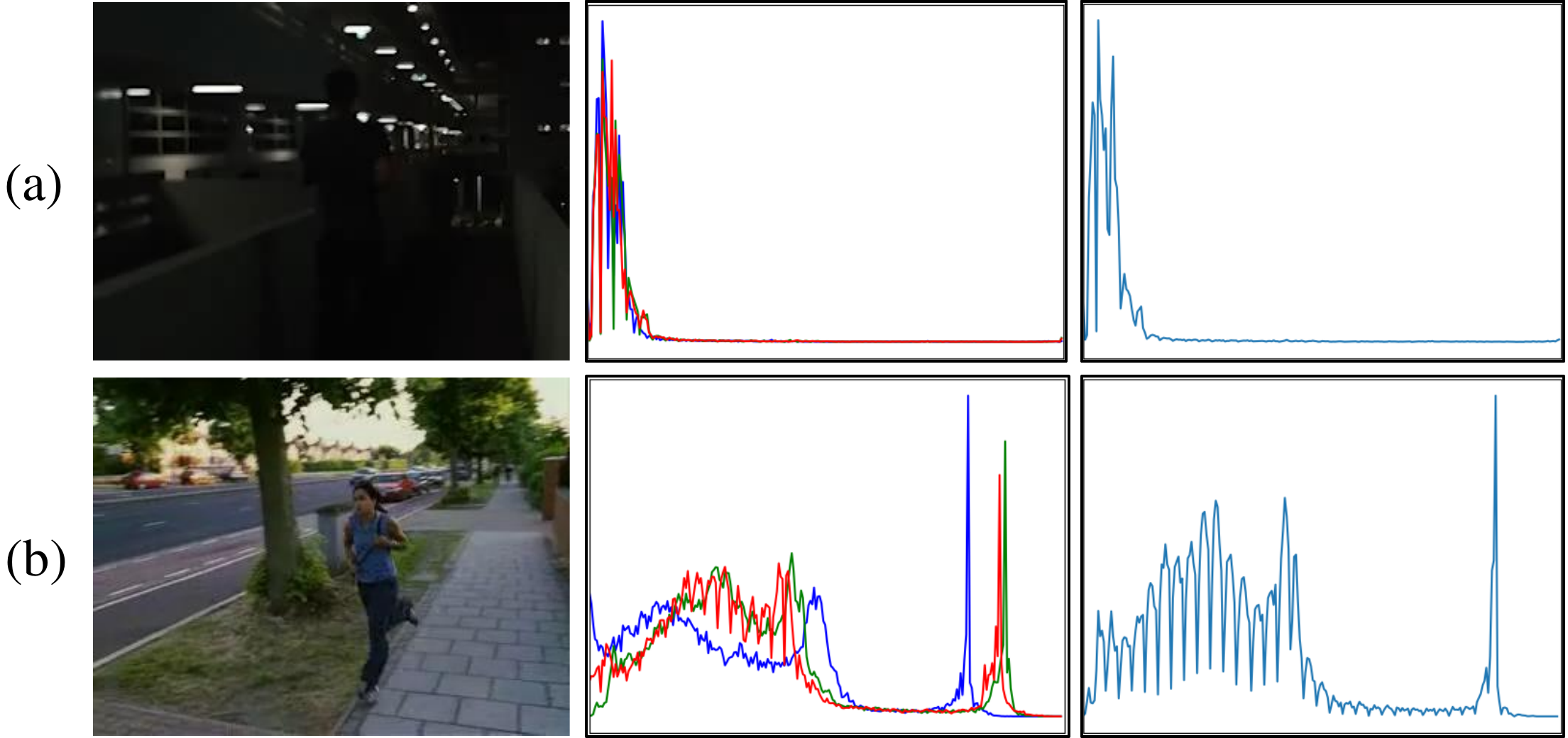}
		\caption{Comparison of sampled frames and the RGB and Y value histograms of their corresponding videos from (a) ARID dataset and (b) HMDB51 dataset. The RGB (middle) and Y value (right) histograms of the video from the ARID dataset are more concentrated at the lower value. Best viewed in color and zoomed in.}
		\label{figure:exp-6-arid_hmdb}
	\end{figure}

	The histograms of ARID, as shown in Figure~\ref{figure:exp-5-rgby_histograms}(a), depict the characteristics of videos in our ARID dataset. Compared to the original HMDB51, the distribution of RGB values in the ARID dataset is much more concentrated towards the region of lower values. Though background light with higher values can be seen in a small portion of videos, they consist of only a small part of the whole frame, thus have few effects on the overall histogram. The fact that pixels in ARID possess lower RGB mean and standard deviation values as shown in Figure~\ref{figure:exp-4-mean_stds} implies that that video frames in ARID are lower in brightness and contrast compare to video frames in HMDB51. This is further justified by the sampled frames and their RGB and Y histograms comparison between ARID and HMDB51 datasets, presented in Figure~\ref{figure:exp-6-arid_hmdb}. The lower brightness and lower contrast for video frames in ARID make it challenging even for the human naked eye to identify actors and the actions in each frame.
	
	We observe that our real dark dataset ARID and the synthetic dark dataset HMDB51-dark are very similar in terms of the Y value, which reflects the luminance of the video frames. This in part shows that our synthesized operation simulates the real dark environment relatively well in terms of video brightness. However, further comparison in terms of RGB values indicates that the real dark dataset ARID is still lower in both RGB mean and standard deviation values. We notice that although the Y values of both ARID and HMDB51-dark are concentrated in the lower values, the histogram of Y values for HMDB51-dark is more spread out. This matches the observation of the RGB histograms where we notice that there are quite a number of pixels in the HMDB51-dark dataset with relatively high pixel values. This is due to the fact that bright pixels exist in the original HMDB51 dataset, which corresponds to the bright background of the frames.
	
\begin{figure}[!t]
		\centering
		\includegraphics[width=.9\linewidth]{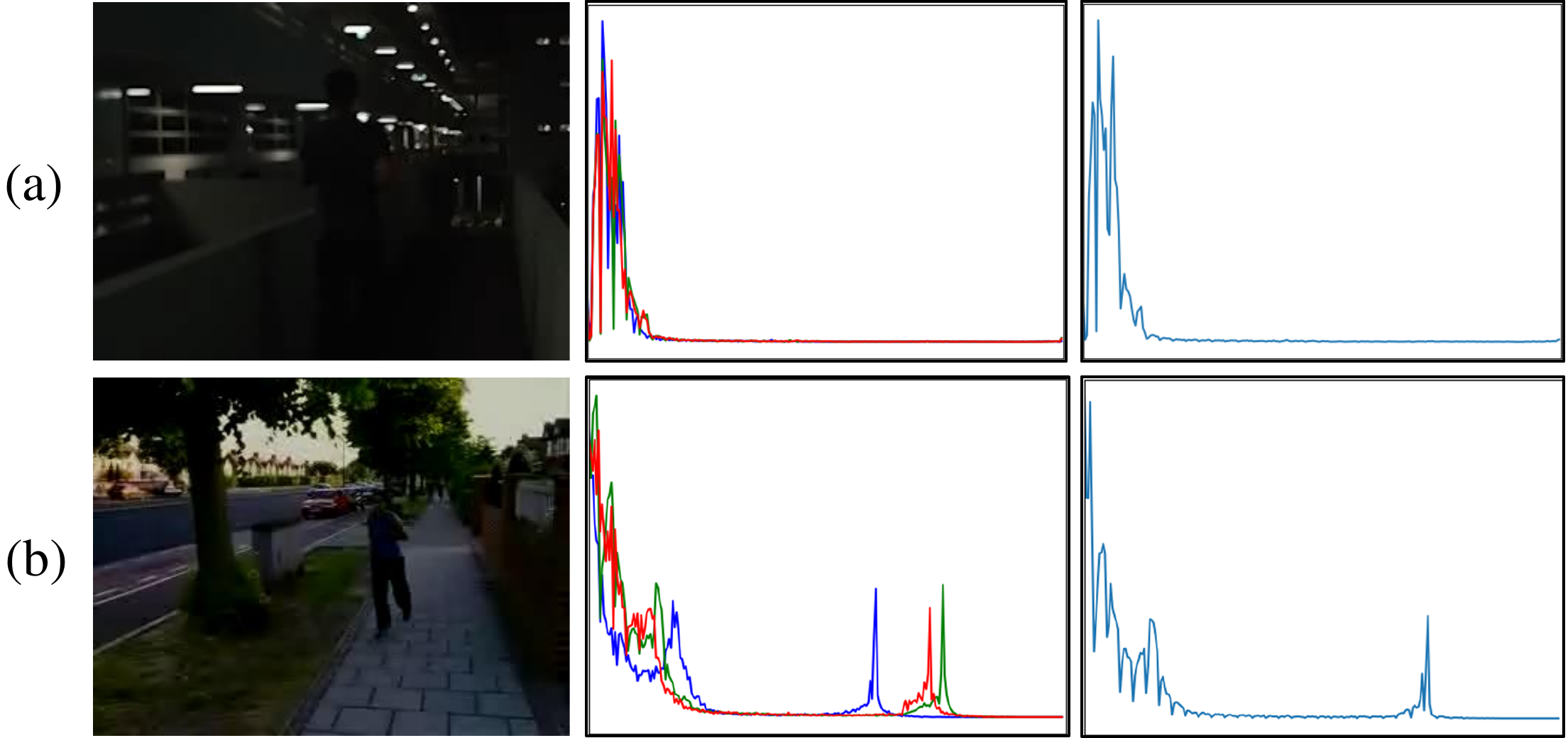}
		\caption{Comparison of sampled frames and the RGB and Y value histograms of their corresponding videos from (a) ARID dataset and (b) HMDB51-dark dataset. The Y value histogram (right) of the HMDB51-dark video is similar to that of the Y value histogram of the ARID video. However, the RGB histogram (middle) of the video from the ARID dataset is still more concentrated. The peaks of the RGB histogram of the HMDB51-dark video comes from the bright background. Best viewed in color and zoomed in.}
		\label{figure:exp-7-arid_hmdbd}
	\end{figure}

	For extreme bright pixels with a normalized value of 1.0, the output synthetic dark pixel value $D_{1.0}(t_{1.0}, x_{1.0}, y_{1.0})$ corresponding to these pixels is still 1.0 according to Equation~\ref{eqn:exp:gamma-darken}. Pixels with high values make up a considerable portion of the overall pixels, as shown in Figure~\ref{figure:exp-5-rgby_histograms}(c), reaching a small peak near the normalized value of 1.0. Hence the corresponding output pixels in the synthetic dark videos have higher pixel values, which raises both the mean value and standard deviation of HMDB51-dark, which in terms is reflected as frames with higher brightness and contrast. This indicates that videos from HMDB51-dark would visually be more distinguishable. The sampled frames and their corresponding RGB and Y histograms comparison between ARID and HMDB51-dark datasets, as shown in Figure~\ref{figure:exp-7-arid_hmdbd} justifies the observation of lower brightness and contrast for video frames in ARID.
	
\begin{figure}[!t]
		\centering
		\includegraphics[width=.9\linewidth]{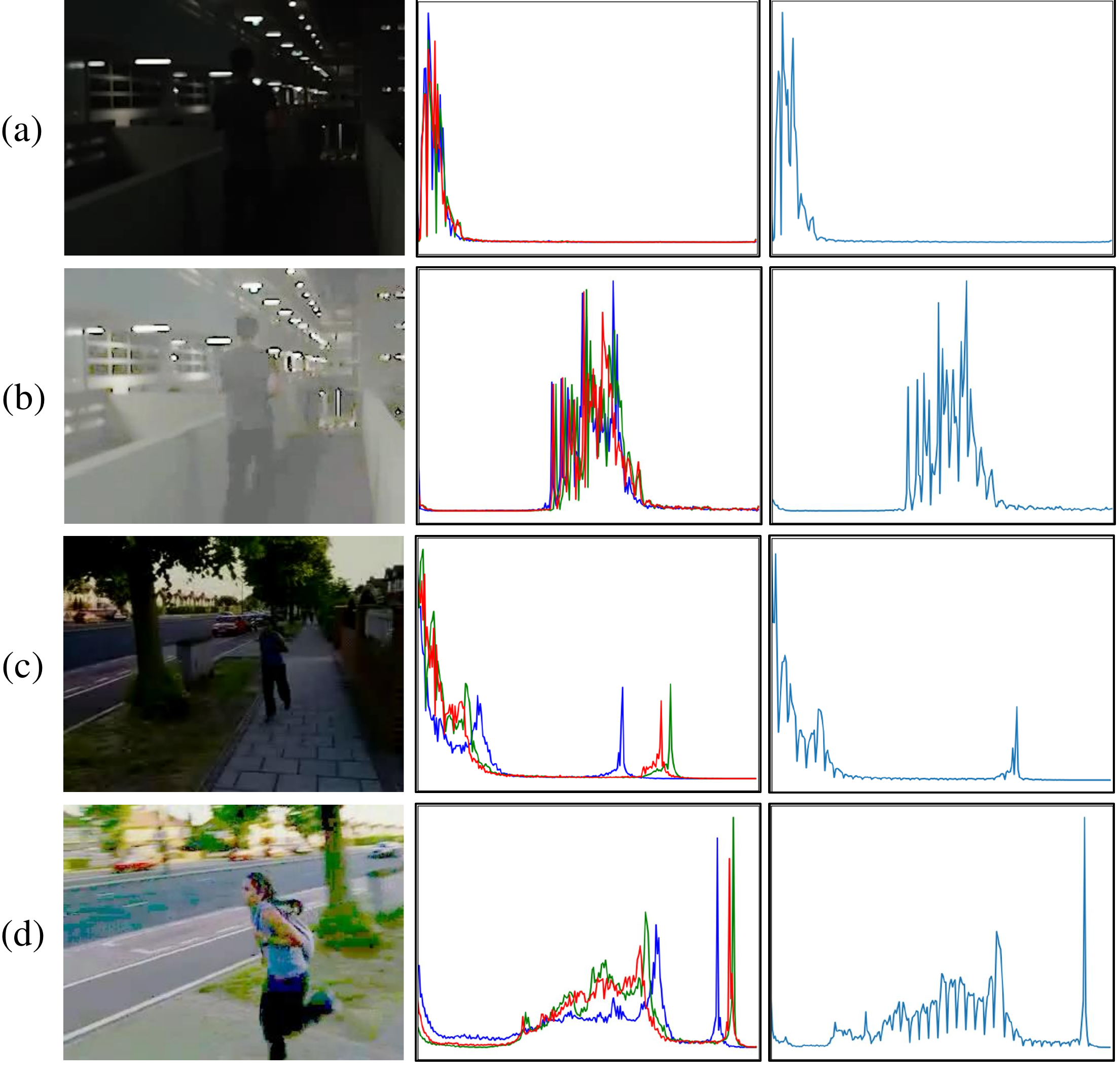}
		\caption{Comparison of sampled frames and the RGB (middle column) and Y (right column) value histograms of their corresponding videos from (a) ARID dataset, (b) ARID-GIC dataset, (c) HMDB51-dark dataset and (d) HMDB51-dark-GIC dataset. \textbf{GIC} enhancement shifts the RGB and Y value histograms towards the larger values, indicating brighter video frames. The RGB and Y values of ARID-GIC are both more concentrated than that of HMDB51-dark-GIC, which matches the low contrast and pale image as shown in the left column. Best viewed in color and zoomed in.}
		\label{figure:exp-8-aridb_hmdbdb}
	\end{figure}

	%%%%%%%%%% 应该利用YCbCr进行说明：即Y值和RGB值的提升以及UV值上提升佐证了GIC对每帧在亮度上和饱和度上提升 %%%%%%%%%%
	As mentioned in Section~\ref{section:exp:frame-enhance}, the \textbf{GIC} method could enhance frames by adjusting the luminance of the frames. By setting $\gamma\geq1.0$, the resulting pixel value after applying the \textbf{GIC} method should be larger than the input pixel value. This is justified through the comparison between Figure~\ref{figure:exp-5-rgby_histograms}(a) and (b), as well as the comparison between Figure~\ref{figure:exp-5-rgby_histograms}(d) and (e). In both cases, the RGB histogram of the dataset after applying the \textbf{GIC} method shows that pixel values would shift towards regions of larger values quite significantly. 
	This is supported by the fact that RGB mean values for both cases increase. Sampled frames as shown in Figure~\ref{figure:exp-8-aridb_hmdbdb} also justifies that \textbf{GIC} enhancement greatly increases the visibility of each video frame. Note that Figure~\ref{figure:exp-8-aridb_hmdbdb}(b) is the direct output after the \textbf{GIC} enhancement of Figure~\ref{figure:exp-8-aridb_hmdbdb}(a). The person seen running can not be clearly observed by the naked eye in Figure~\ref{figure:exp-8-aridb_hmdbdb}(a), whereas the person becomes more visible in Figure~\ref{figure:exp-8-aridb_hmdbdb}(b). The brighter and clearer video frames also further justify the comparison of the Y histograms, where the Y histograms for videos after applying \textbf{GIC} method would also shift to larger values, indicating a rise in luminance for video frames.
	
	%%%%%%%%%% 可能应该加上利用YCbCr进行说明：即UV值和RGB值上std的区别对contrast的体现 %%%%%%%%%%
	Though both the histograms and sampled frames of ARID-GIC and those of HMDB51-dark-GIC show the effectiveness of \textbf{GIC} enhancement in increasing the luminance of dark videos, there are still significant differences between the histograms of either dataset. The most significant difference is that pixel values of ARID-GIC are much more concentrated compared with pixel values of HMDB51-dark-GIC. This is in line with the fact that pixel values of ARID are concentrated compared with pixel values of HMDB51-dark. Unlike the \textbf{HE} enhancement where the distribution of pixel values could be altered and creates higher contrast video frames, the \textbf{GIC} enhancement focuses on adjusting the illumination of video frames. Hence the characteristic of pixel value distribution would not change drastically after the \textbf{GIC} enhancement. The concentration of pixel values indicates that video frames in ARID-GIC are still low in contrast. This is further justified by comparing the sampled frames as shown in Figure~\ref{figure:exp-8-aridb_hmdbdb}(b) and (d). The sampled frame from ARID-GIC looks pale as compared to the sampled frame from HMDB51-dark-GIC as shown in Figure~\ref{figure:exp-8-aridb_hmdbdb}(b).
	
	From the above observation, we can summarize the main characteristic of the real dark videos collected in our ARID dataset: low brightness and low contrast. Though the character of low brightness could be simulated by the synthetic dark videos, the characteristic of low contrast cannot be easily replicated by synthetic dark videos. This is partly due to the bright backgrounds and pixels that commonly exist in videos shot under normal illumination. The fact that synthetic dark videos are higher in contrast compared to real dark videos confirms that real dark videos are necessary and irreplaceable for the task of action recognition in a dark environment.
	
	\subsection{Classification Results on ARID}
	\label{section:exp:classify-analysis}
	In this section, we illustrate how current action recognition models perform in the task of action recognition in the dark on our ARID dataset. We further explore potential ways to improve the performance of action recognition in real dark videos and reveal some challenges faced with action recognition in dark videos. 
	
	\subsubsection{Model Benchmarking}
	The performances of current action recognition models are presented in Table~\ref{table:exp-1-benchmark_arid}, which includes: 
	\begin{itemize}
		\item \textbf{Two-stream models: }the original two-stream method with VGG~\cite{simonyan2014very} backbone (VGG-TS)~\cite{simonyan2014two}, Temporal Segment Network (TSN)~\cite{wang2016temporal} and I3D with two-stream inputs (I3D-TS)~\cite{carreira2017quo}; 
		\item \textbf{3D-CNN based models: }C3D~\cite{tran2015learning}, Separable-3D~\cite{xie2018rethinking}, 3D-ShuffleNet~\cite{kopuklu2019resource}, 3D-SqueezeNet~\cite{iandola2016squeezenet}, 3D-ResNet-18~\cite{tran2018closer}, 3D-ResNet-50~\cite{hara2018can}, 3D-ResNet-101~\cite{hara2018can}, I3D with RGB input (I3D-RGB)~\cite{carreira2017quo}, Pseudo-3D-199~\cite{qiu2017learning} and 3D-ResNext-101~\cite{hara2018can}.
	\end{itemize}
	
	\begin{table}[!t]
		\centering
		\resizebox{1.\linewidth}{!}{
			\begin{tabular}{c|c|cc}  
				\hline
				\hline
				& Models          & Top-1 Accuracy    & Top-5 Accuracy \\
				\hline
				\multirow{3}{*}{Two-stream}
				& VGG-TS          & 32.08\%           & 90.76\% \\
				& TSN             & 57.96\%           & 94.17\% \\
				& I3D-TS          & 72.78\%           & 99.39\% \\
				\hline
				\multirow{10}{*}{3D-CNN}
				& C3D             & 40.34\%           & 94.17\% \\
				& Separable-3D    & 42.16\%           & 93.44\% \\
				& 3D-ShuffleNet   & 44.35\%           & 93.44\% \\
				& 3D-SqueezeNet   & 50.18\%           & 94.17\% \\
				& 3D-ResNet-18    & 54.68\%           & 96.60\% \\
				& I3D-RGB         & 68.29\%           & 97.69\% \\
				& 3D-ResNet-50    & 71.08\%           & 99.39\% \\
				& 3D-ResNet-101   & 71.57\%           & 99.03\% \\
				& Pseudo-3D-199   & 71.93\%           & 98.66\% \\
				& 3D-ResNext-101  & 74.73\%           & 98.54\% \\
				\hline
				\hline
			\end{tabular}
		}
		\caption{Performance of current two-stream and 3D-CNN based action recognition models on ARID dataset.}
		\label{table:exp-1-benchmark_arid}
	\end{table}
	
	The performance results in Table~\ref{table:exp-1-benchmark_arid} show that among the current two-stream and 3D-CNN-based action recognition models examined, 3D-ResNext-101 achieves the best performance with a top-1 accuracy of $74.73\%$. We notice that the top-5 accuracies are relatively high for all methods, partly because of the limited number of classes in our dataset. By comparing the performance of I3D with only RGB input and with two-stream inputs, a noticeable increase of $4.49\%$ in accuracy is observed by adding optical flow as the temporal input. This increase highly suggests that optical flow could provide a more accurate and effective representation of videos in our ARID dataset. However, such an increase comes at a cost of pre-computing optical flow, which is expensive in both computation power and storage. Hence we focus on 3D-CNN-based models for the rest of our paper.
	
	We also notice that though our dataset is of relatively small size and has fewer classes than current normal illumination video datasets, there is plenty of room for improvement in accuracy. To explore potential ways for further improving accuracy for dark videos, we selected 3D-CNN-based models, i.e.\ C3D, I3D-RGB, 3D-ResNet-101, and 3D-ResNext-101, as the baselines for further experiments. 
	
	\subsubsection{Improvement Exploration with Synthetic Dark Dataset}
	An intuitive method for improving accuracy is by using frame enhancement methods as introduced in Section~\ref{section:exp:frame-enhance}. To test whether frame enhancement methods could improve accuracy, we employ the \textbf{GIC} method on the synthetic HMDB51-dark dataset due to its larger data size and ease of obtaining dark data from the current datasets. The performance of the chosen 3D-CNN-based models on the synthetic dataset HMDB51-dark and its corresponding \textbf{GIC} enhanced HMDB51-dark-GIC is illustrated in Table~\ref{table:exp-2-frame_enhance_hmdb}.
	
	\begin{table}[t]
		\centering
		\resizebox{1.\linewidth}{!}{
			\begin{tabular}{l|C{2cm}C{2cm}C{2cm}}  
				\hline
				\hline
				Models & HMDB51-dark & HMDB51-dark-GIC & HMDB51\\
				\hline
				C3D                 & 21.13\% & 21.64\% & 50.13\%\\
				I3D-RGB             & 27.90\% & 41.79\% & 54.64\%\\
				3D-ResNet-101       & 42.48\% & 50.78\% & 61.70\% \\
				3D-ResNext-101      & 44.90\% & 58.62\% & 63.80\% \\
				\hline
				\hline
			\end{tabular}
		}
		\caption{Performance of various 3D-CNN based action recognition models on the synthetic HMDB51-dark and its \textbf{GIC} enhanced HMDB51-dark-GIC. The performance of the respective models on the original HMDB51 is presented for reference.}
		\label{table:exp-2-frame_enhance_hmdb}
	\end{table}
	
	The results in Table~\ref{table:exp-2-frame_enhance_hmdb} exhibit sharp decreases in classification accuracies when the same networks are utilized for the dark data. The decreases in accuracies are expected, given that dark videos contain fewer details as displayed in Figure~\ref{figure:exp-3-hmdb_hmdbd}, hence less information could be extracted for recognizing actions. Besides, we also notice consistent increases in accuracies when the \textbf{GIC} method is applied to enhance the dark video frames. The degrees of increases vary across different models. However, besides the C3D model, the degrees of increases for the other three models all exceed $8\%$, which is rather significant. The largest degree of increase of $13.89\%$ is achieved with the I3D-RGB model. As the synthetic data is darkened with random gamma values while the \textbf{GIC} enhancement utilizes a fixed gamma value, it is nearly impossible to recover the original videos. Despite this, our results show that the \textbf{GIC} enhancement still brings a relatively consistent and significant amount of accuracy improvements for most models through enhancing each video frame.
	
	\subsubsection{Applying Frame Enhancement Methods to ARID}
	The success in applying the straightforward frame enhancement method of \textbf{GIC} in increasing classification accuracies for synthetic dark videos gives us a hint on potential ways to improve accuracy for action recognition in real dark videos. To justify if the same \textbf{GIC} method could also improve action recognition accuracy on our ARID dataset, we perform experiments on the \textbf{GIC} enhanced ARID dataset: ARID-GIC, utilizing the four 3D-CNN based models aforementioned. The results are as presented in Table~\ref{table:exp-3-multi_enhance_arid} with the Top-1 accuracy and the relative improvements compared to their respective performances on the original ARID dataset.
	
	\begin{table*}[!t]
		\centering
		\resizebox{.7\linewidth}{!}{
			\begin{tabular}{l|l|C{2cm}C{2cm}C{2cm}C{2cm}}  
				\hline
				\hline
				Datasets    & Accuracy & C3D & I3D-RGB & 3D-ResNet-101 & 3D-ResNext-101 \\
				\hline
				\multirow{2}{*}{ARID-GIC}
				& Top-1     & 44.09\% & 69.14\% & 75.15\% & 78.06\% \\
				& Improv.   & 3.75\% & 0.85\% & 3.58\% & 3.33\% \\
				\hline
				\multirow{2}{*}{ARID-HE}
				& Top-1     & 39.49\% & 63.67\% & 65.49\% & 75.82\% \\
				& Improv.   & -0.85\% & -4.62\% & -6.08\% & 1.09\% \\
				\hline
				\multirow{2}{*}{ARID-LIME}
				& Top-1     & 39.61\% & 73.02\% & 75.45\% & 77.40\% \\
				& Improv.   & -0.73\% & 4.73\% & 3.88\% & 2.67\% \\
				\hline
				\multirow{2}{*}{ARID-BIMEF}
				& Top-1     & 45.23\% & 68.89\% & 68.28\% & 73.39\% \\
				& Improv.   & 4.89\% & 0.60\% & -3.29\% & -1.34\% \\
				\hline
				\multirow{2}{*}{ARID-KinD}
				& Top-1     & 46.64\% & 67.55\% & 70.59\% & 69.62\% \\
				& Improv.   & 6.30\% & -0.74\% & -0.98\% & -5.11\% \\
				\hline
				ARID        & Top-1 & 40.34\% & 68.29\% & 71.57\% & 74.73\% \\
				\hline
				\hline
			\end{tabular}
		}
		\caption{Performance of various 3D-CNN based action recognition models on variants of ARID enhanced by \textbf{GIC},  \textbf{HE}, \textbf{LIME}, \textbf{BIMEF} and \textbf{KinD}. The Improvements (Improv.) are compared with the performances of the respective models on the original ARID dataset, which is also presented for reference.}
		\label{table:exp-3-multi_enhance_arid}
	\end{table*}
	
	The results in Table~\ref{table:exp-3-multi_enhance_arid} illustrate that the action recognition accuracies on the ARID dataset could be improved consistently through \textbf{GIC} enhancement with all models, thanks to the increase in the illumination of each video frame as presented in Figure~\ref{figure:exp-8-aridb_hmdbdb}. The increase in accuracy is consistent with findings in the synthetic dark dataset HMDB51-dark. However, we also notice that the improvements of performances by using \textbf{GIC} on ARID are rather limited compared to the improvements in the synthetic dark dataset. The degrees of improvements are capped at $3.75\%$, while three out of the four models experience less improvement than that in the synthetic HMDB51-dark. As \textbf{GIC} method is a straightforward method based on simple exponential calculation, we further examine if more sophisticated frame enhancement methods could further improve action recognition accuracy. We thus examine the accuracy on datasets ARID-HE, ARID-LIME, ARID-BIMEF, and ARID-KinD, which are results of the output by frame enhancement methods \textbf{HE}, \textbf{LIME}, \textbf{BIMEF} and \textbf{KinD} respectively using the same models. The results are presented in Table~\ref{table:exp-3-multi_enhance_arid}.
	
	Interestingly, Table~\ref{table:exp-3-multi_enhance_arid} demonstrates that not all frame enhancement methods result in improvements in action recognition accuracies for dark videos. Of all the frame enhancement methods, the most consistent improvement is achieved by the rather simple \textbf{GIC} method. Whereas the accuracy drops for most networks when utilizing the recent deep learning-based method \textbf{KinD} and the \textbf{HE} method. We also observe that none of the improvements matches that achieved in the \textbf{GIC}-enhanced HMDB51-dark-GIC dataset. 
	
	\subsubsection{Effects of Frame Enhancement Methods on ARID}
\begin{figure}[!t]
		\centering
		\includegraphics[width=1.0\linewidth]{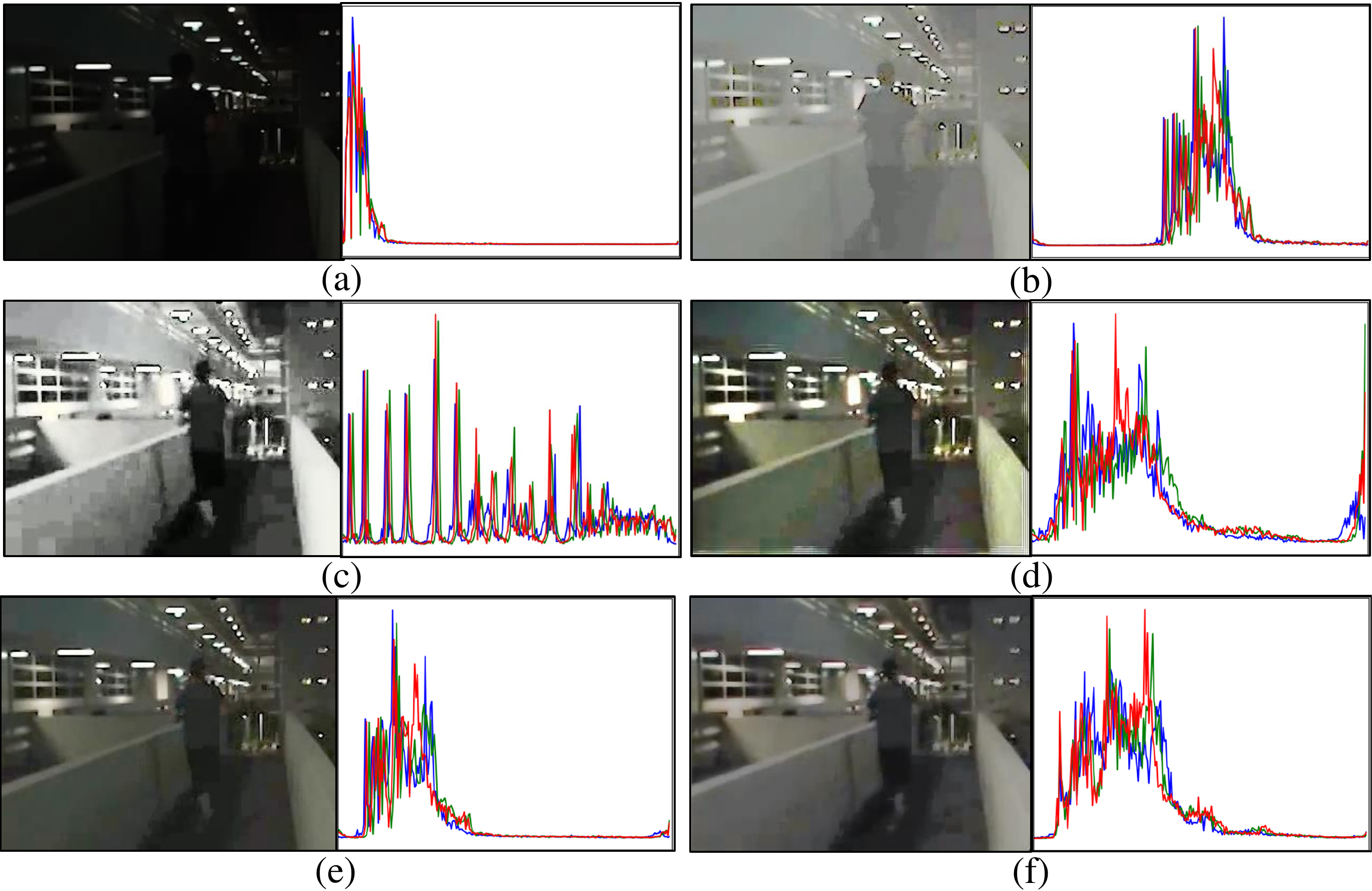}
		\caption{Comparison of the sampled frames and their respective RGB histograms from (a) ARID, (b) ARID-GIC, (c) ARID-HE, (d) ARID-LIME, (e) ARID-BIMEF and (f) ARID-KinD. Best viewed in color.}
		\label{figure:exp-9-arid_multi_b_hist_all}
	\end{figure}

	To gain a better understanding of the difference between the outcome of utilizing the different enhancement methods, we visualize the frame output of each enhancement method. Figure~\ref{figure:exp-9-arid_multi_b_hist_all} presents the sampled frames and their respective RGB histograms of the output of the above enhancement methods with the same input ARID video frame.
	
	Figure~\ref{figure:exp-9-arid_multi_b_hist_all} clearly shows that visually, the outputs of all frame enhancement methods improve the visibility of the video. The actor who is running can be seen clearly in all sampled frames except the sample frame from the original video in ARID substantially. In fact, the sampled frame of ARID-GIC does not appear to be the best enhancement visually, as it is still low in contrast.
	
	In comparison, all other methods produce higher contrast images, as justified by the RGB histograms in Figure~\ref{figure:exp-9-arid_multi_b_hist_all}. This indicates that current frame enhancement which clearly improves dark video frames visually may not bring improvement in action recognition accuracy for dark videos. We argue that some enhancements can be regarded as artifact or adversarial attacks for videos. Though enhanced frames are clearer visually, some enhancements break the original distribution of videos and introduce noise. The change in distribution and introduction of noise could lead to a decrease in performance for action recognition models.
	
	\subsubsection{Accuracy for Each Action Class}
\begin{figure*}[!t]
		\centering
		\includegraphics[width=.75\linewidth]{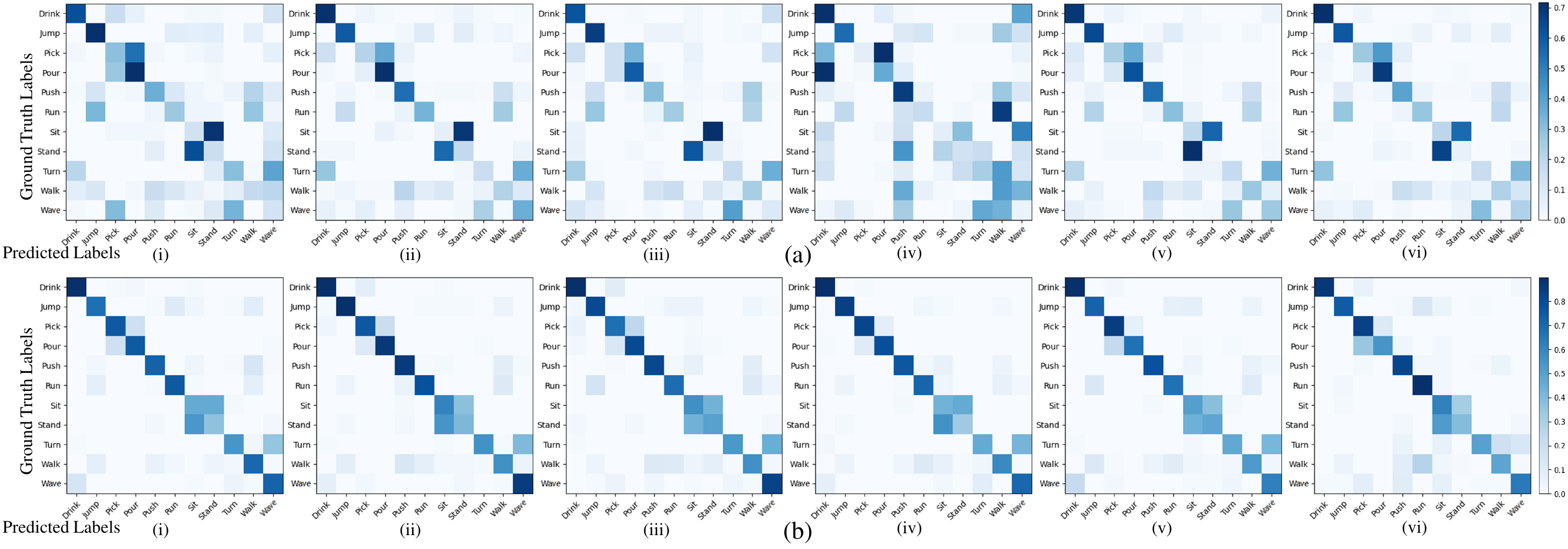}
		\caption{Comparison of the normalized confusion matrices for (a) C3D and (b) 3D-ResNext-101 models. The normalized confusion matrices show the accuracies for each class as the values at the diagonal corresponding to the ground truth labels at the vertical axis. The normalized confusion matrices are constructed with respect to (i) ARID, (ii) ARID-GIC, (iii) ARID-HE, (iv) ARID-LIME, (v) ARID-BIMEF and (vi) ARID-KinD. Best viewed in color and zoomed in.}
		\label{figure:exp-10-cm_arid}
	\end{figure*}

	To further understand how the action recognition performs for each individual class, we present the confusion matrices for the ARID dataset and all its variants. In particular, the confusion matrices in are constructed using the worst performing model, C3D model (Figure~\ref{figure:exp-10-cm_arid}(a)), and the best performing model, 3D-ResNext-101 model (Figure~\ref{figure:exp-10-cm_arid}(b)) respectively.
	
	The pattern in Figure~\ref{figure:exp-10-cm_arid} clearly presents the performance of the models for the different classes with different frame enhancement methods. In general, the patterns of the confusion matrices for the same model remain largely the same for the original ARID dataset and all of its variants. This is in line with the results shown in Table~\ref{table:exp-3-multi_enhance_arid} where none of the frame enhancement methods would bring a significant increase in accuracy. In most cases, four of the eleven classes could reach an accuracy of greater than $50\%$: Drinking, Jumping, Pouring, and Pushing. We observe that for these classes, the visible background would change more obviously due to the action. For example, the action "Jumping`` could cause light in the background to be seen flashing as the actor jumps up and down.
	
	It could further be observed that a large improvement is achieved by utilizing a much deeper network as 3D-ResNext-101. However, there are still several classes in which most videos are misclassified. One noticeable case is for classes "Standing`` and "Sitting``, where relatively large percentages of either class being misclassified as the other class. The two classes in our dataset share similar action patterns and differ in the order sequence of frames. The other noticeable case is for classes "Turning`` and "Waving``, where they differ mostly in the movement of the hand.
	
	\subsection{Feature Visualization with ARID}
	\label{section:exp:visualization}
	\begin{figure}[!t]
		\centering
		\includegraphics[width=.9\linewidth]{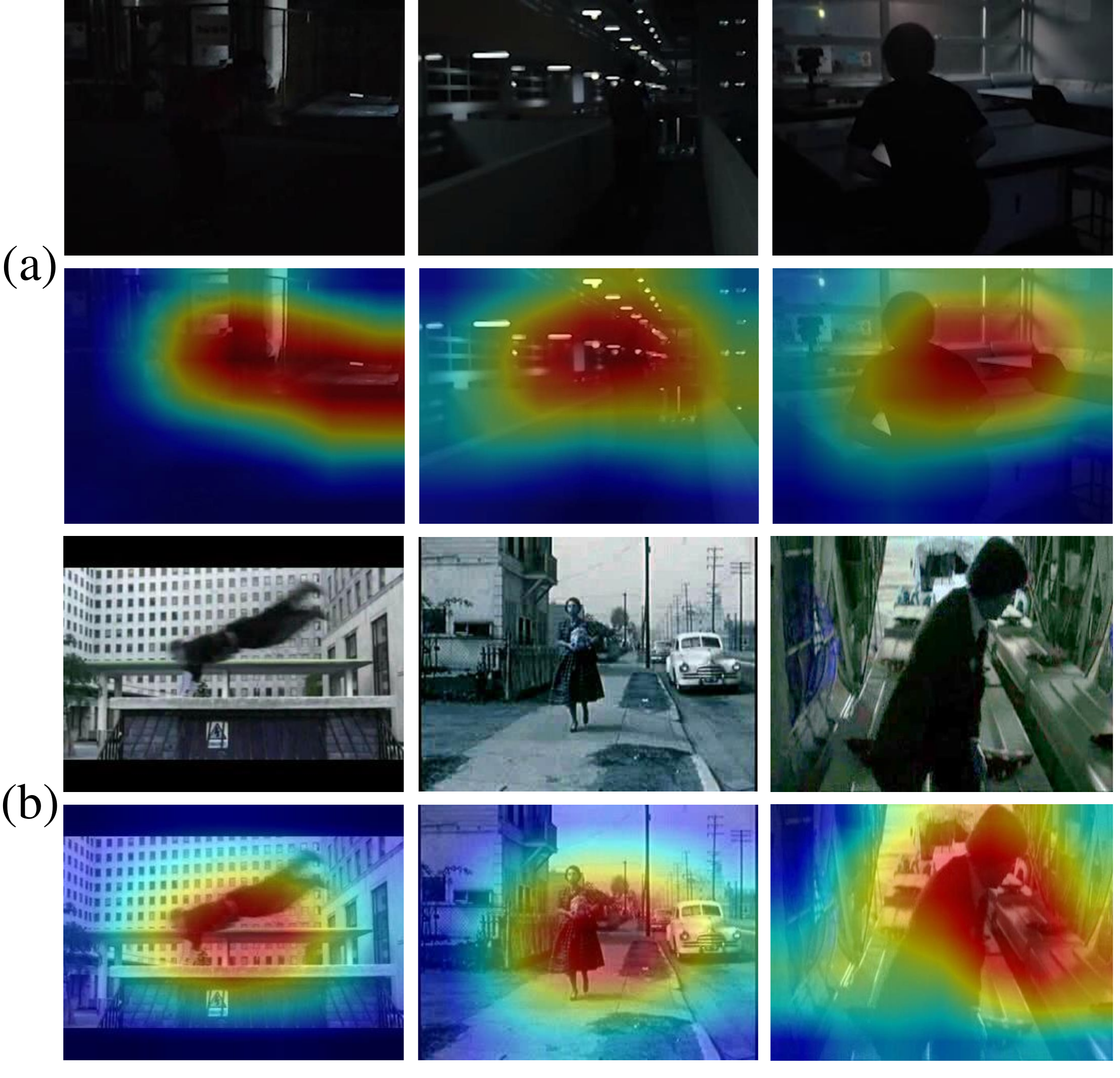}
		\caption{Comparison of sampled frames and their corresponding \textit{CAM}s from (a) ARID and (b) HMDB51 dataset, extracted by utilizing 3D-ResNext-101 model. We present sampled frames from three common classes: Jumping (left), Running (mid) and Standing (right). Best viewed in color and zoomed in.}
		\label{figure:exp-11-cam_arid_hmdb}
	\end{figure}
	
	To further understand the performance of current action recognition models on ARID and analyze the effect of dark videos on current models, we extract and visualize features at the last convolution layer of the action recognition models. The visualizations of features are presented as \textit{Class Activation Maps (CAM)}~\cite{zhou2016learning}, which depicts the focus of the model with respect to the given prediction. Specifically, our \textit{CAM}s are extracted by utilizing the 3D-ResNext-101 model first, due to the best performance achieved by the 3D-ResNext-101 model on both HMDB51 and ARID datasets. Figure~\ref{figure:exp-11-cam_arid_hmdb} compares the sampled frames from the ARID and HMDB51 datasets, with the corresponding \textit{CAM}s extracted by 3D-ResNext-101 trained on either datasets. We observe that for the frames in HMDB51 with normal illumination, the 3D-ResNext-101 model is able to focus on the actors. The model is able to explicitly exclude most of the information from the background. Whereas for the dark video, instead of focusing on the actor, the model focuses more on the surroundings. For example, for the action in Figure~\ref{figure:exp-11-cam_arid_hmdb}(a)(left), the network classifies the action as "Jumping`` not by focusing on the person. Rather it focuses on the background whose details are uncovered due to the person jumping backward. Therefore the \textit{CAM} shows the network focusing on a narrow beam in the background. The focus on the background instead of the actor could be partly due to the fact that clear outlines of actors rarely exist in dark videos.
	
\begin{figure*}[!t]
		\centering
		\includegraphics[width=.75\linewidth]{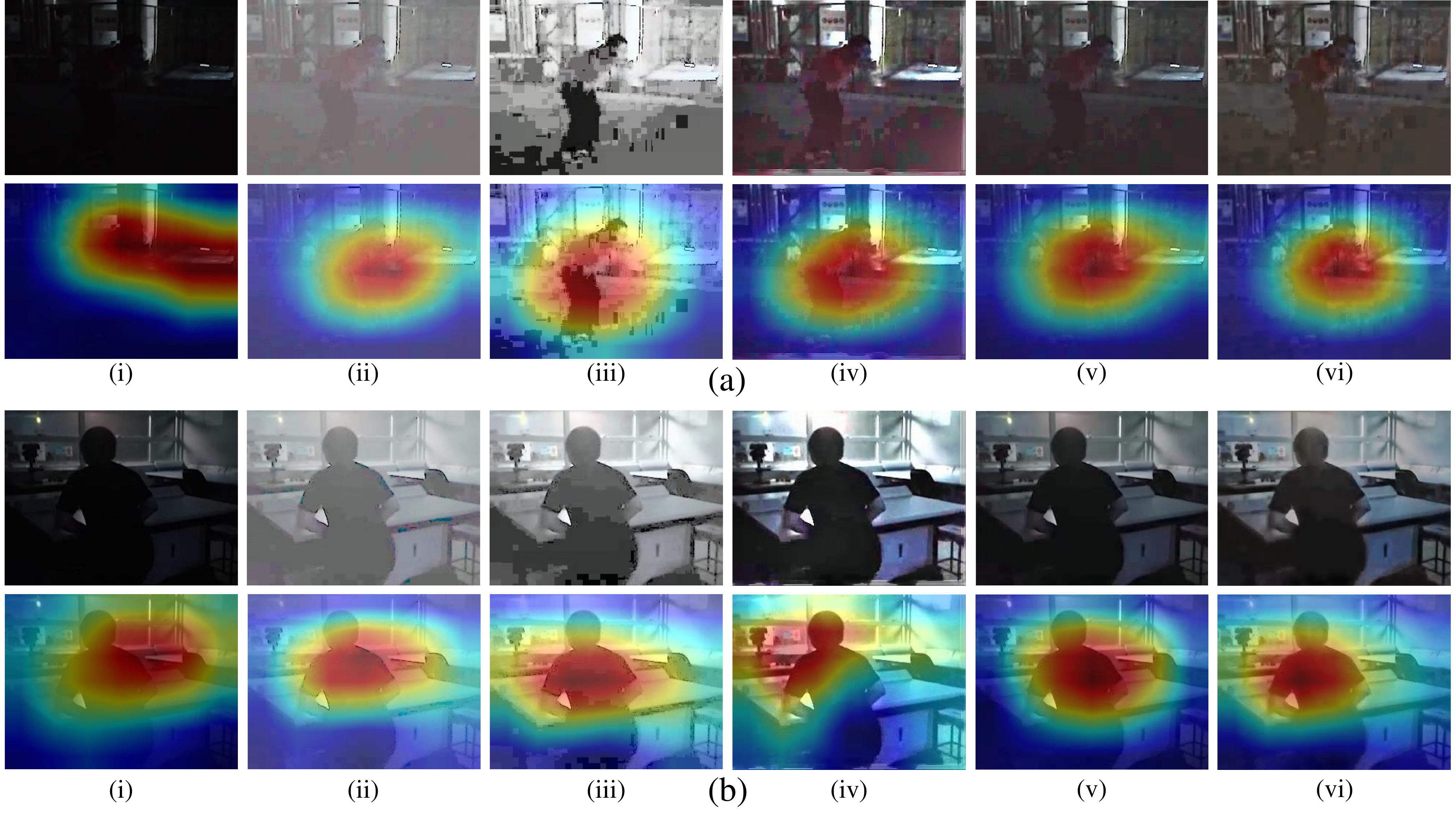}
		\caption{Comparison of sampled frames and their corresponding \textit{CAM}s of classes: (a) Jumping and (b) Standing, extracted by utilizing 3D-ResNext-101 model. The sampled frames and their \textit{CAM}s are from (i) ARID, (ii) ARID-GIC, (iii) ARID-HE, (iv) ARID-LIME, (v) ARID-BIMEF and (vi) ARID-KinD.}
		\label{figure:exp-12-cam_arid_multi_b}
	\end{figure*}

	From Table~\ref{table:exp-3-multi_enhance_arid}, we have concluded that for each model, certain frame enhancement methods could positively affect the final classification accuracy. To gain further understanding of how the different frame enhancement methods actually affect the 3D-ResNext-101 model, we compare the \textit{CAM}s with respect to the same sampled frame from the original ARID dataset and the five frame enhanced ARID datasets as shown in Figure~\ref{figure:exp-12-cam_arid_multi_b}. Compared to the original video frame, the outline of the actor is much clearer in all enhanced frames. The focus area of the network is significantly more concentrated towards the actor compared with \textit{CAM} of the original frame, where it includes large portions of the background. However, noticeable offsets do exist between the focus of the network of the frame enhanced sample frames and the actual actor. The offsets show that the focus area of networks would still include portions of the background, which is rather irrelevant to the actor or the action. In contrast, the \textit{CAM}s of HMDB51 video frames show the areas of network focus center explicitly around the actors. This may partly explain the inability of frame enhancement methods to improve action recognition accuracy while being able to focus on a more concentrated area of each video frame.
	
\begin{figure*}[!t]
		\centering
		\includegraphics[width=.75\linewidth]{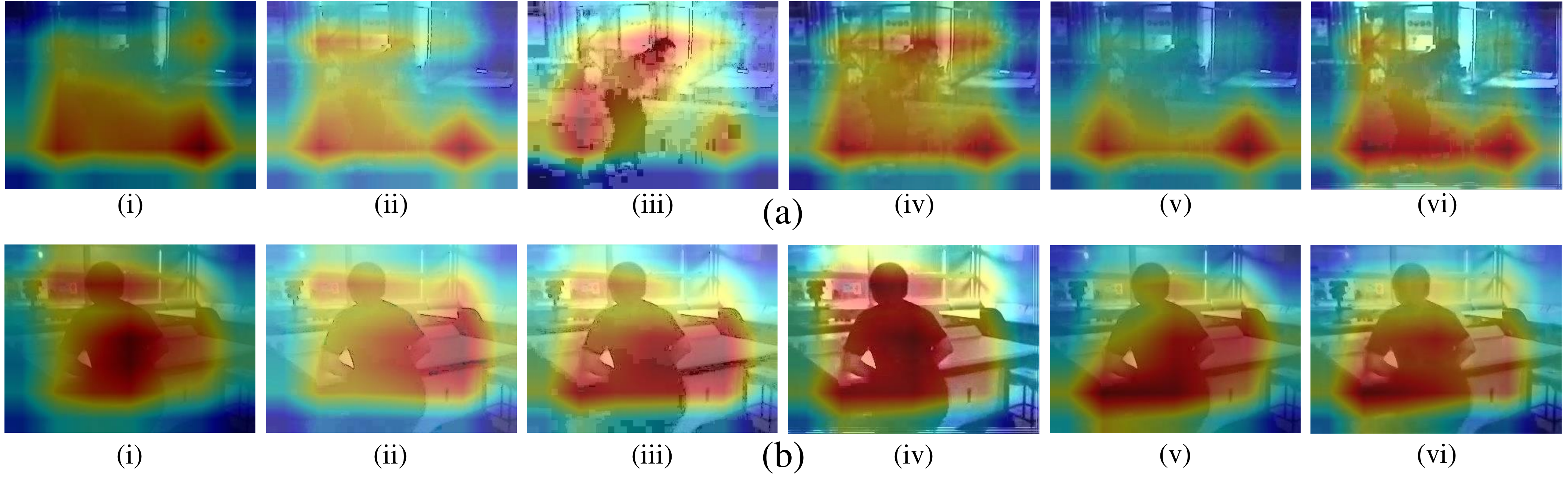}
		\caption{Comparison of the \textit{CAM}s of classes: (a) Jumping and (b) Standing, extracted by utilizing C3D model, corresponding to the same sampled frames in Figure~\ref{figure:exp-12-cam_arid_multi_b}. The \textit{CAM}s are extracted from (i) ARID, (ii) ARID-GIC, (iii) ARID-HE, (iv) ARID-LIME, (v) ARID-BIMEF and (vi) ARID-KinD datasets.}
		\label{figure:exp-13-cam_c3d_arid_multi_b}
	\end{figure*}

	We notice that the effect of the same frame enhancement methods varies across different action recognition models. Among the five frame enhancement methods, four of them produce opposite effects on C3D and 3D-ResNext-101 models. To further understand how the frame enhancement methods affect the different models, the \textit{CAM}s extract by C3D is presented as shown in Figure~\ref{figure:exp-13-cam_c3d_arid_multi_b} for analysis and comparison. 
	
	The comparison of \textit{CAM}s extracted by the C3D model shows that the concentration effect by applying frame enhancement is unapparent compared to the case where the 3D-ResNext-101 model is utilized. We observe that the focus area for C3D is significantly more spread out, indicating that C3D focuses more on unrelated backgrounds. This explains the lower accuracy of C3D compared to the 3D-ResNext-101 model. By comparison, the dark red region where the network weights the most is relatively more concentrated when \textbf{KinD} or \textbf{BIMEF} methods are used compared to that when \textbf{HE} or \textbf{LIME} methods are used. This coincides with the results as presented in Table~\ref{table:exp-3-multi_enhance_arid}, which suggest that \textbf{KinD} and \textbf{BIMEF} methods produce positive improvements while \textbf{HE} or \textbf{LIME} methods result in negative improvements.
	
	\subsection{Discussion}
	\label{section:exp:summary}
	From the results and analysis presented above, we can draw three major conclusions about our ARID dataset and the task of action recognition in the dark from statistics and classification result aspects:
	\begin{itemize}
		\item Videos taken in a dark environment are characterized by their low brightness and low contrast. Among which the characteristic of low contrast cannot be fully synthesized by synthetic dark videos, therefore synthetic dark videos cannot be directly applied to the task of action recognition in the dark.
		\item Though current frame enhancement methods could produce visually clearer video frames, the accuracy improvements made for current action recognition models after frame enhancing dark videos are rather limited. Some frame enhancement methods even deteriorate classification accuracy, since some enhancements can be regarded as artifacts or adversarial attacks for videos. Breaking the original distribution of videos might decrease the performance of a statistical model. Better frame enhancement methods developed may be helpful in further improving action recognition accuracy in dark videos.
		\item Current action recognition models fail to focus on the actor for classification in many dark videos. This might be caused by unclear outlines of actors and shows that action recognition models could tend to focus on the actors for frame-enhanced dark videos. However, the focuses in frame-enhanced dark videos contain offsets. We believe that better action recognition models with a better ability to focus on actors, especially with unclear outlines, could be a critical part of improving action recognition accuracy in dark videos.
	\end{itemize}
	The above dataset pioneered by us provides fundamental data for the task of action recognition in the dark. The relevant conclusions made through benchmarking current action recognition models provide basic statistical and characteristic analysis of ARID that could contribute to exploring more effective solutions for ARID.

	%------------------------------------------------------------------------
	\section{Conclusion and Future Work}
	\label{section:conclusion}
	
	In this work, we introduced the Action Recognition In the Dark (ARID) dataset, which is, as far as we are aware, the first dataset dedicated to the task of action recognition in the dark. The ARID includes 4k video clips with 11 action categories. To understand the challenge behind real dark videos, we analyze our ARID dataset with three perspectives: statistical, classification result, and feature visualization. We discover distinct characteristics of real dark videos, proving its necessity over synthetic dark videos. Our analysis shows that current action recognition models and frame enhancement methods may not be effective in recognizing action in dark videos. We hope this study could draw more interest from the community to work on the task of action recognition in the dark.
	
	In the future, we would expand our ARID dataset to include more action classes and data, providing a more solid data foundation for the task of action recognition in the dark. Also, we would develop action recognition models that fit the dark environment, improving action recognition accuracy for actions in the dark. Further, we would explore on adapting models trained on synthetic dark datasets to real dark videos through domain adaptation, such that we can make use of current datasets with a vast amount of video data.

	%------------------------------------------------------------------------
	% \clearpage
	\bibliographystyle{named}
	\bibliography{ijcai_full}
	
\end{document}